\documentclass{article}
\usepackage{iclr2027_conference,times}
\usepackage{amsmath,amssymb,mathtools}
\usepackage{booktabs,multirow,makecell}
\usepackage{graphicx}
\usepackage[nopatch=eqnum]{microtype}
\usepackage{xcolor}
\usepackage{hyperref}
\usepackage{url}
\usepackage{float}
\hypersetup{hidelinks,pdftitle={AcrossVAM3.0: Resource-Bounded 4D Scene Construction and Adaptive Surface Association},pdfauthor={Nan Wu and Hong Yee Hung}}

\graphicspath{{figures/}}
\newcommand{\method}{\textsc{AcrossVAM3.0}}
\newcommand{\scene}{\textsc{Scene4D}}
\newcommand{\mhac}{\textsc{MHAC}}

\title{AcrossVAM3.0: Resource-Bounded\\4D Scene Construction and\\Adaptive Surface Association}

\author{Nan Wu\thanks{These authors contributed equally to this work.} \quad
Hong Yee Hung\footnotemark[1]\\
Across Physcis (Beijing) Technology Co., Ltd}

\iclrfinalcopy

\begin{document}
\maketitle
\lhead{AcrossVAM3.0}

\begin{abstract}
Combining video generation, dynamic reconstruction and monocular SLAM requires consistent coordinates, explicit handling of missing outputs and predictable memory use. We present \method{}, an offline system that converts text, images or monocular RGB video into a versioned point representation. Frozen components run sequentially on one GPU; map corrections update cameras, points and trajectories together. The coordination layer adds no neural parameters. Three fresh input-mode runs completed with a maximum sampled task-process-tree GPU footprint of 17.14 GB. In a frozen 16-scene video suite, every scene yielded a 4D reconstruction and 15 yielded a non-empty SLAM map. Strict cross-window matching reduced local surface error but lost long-range track coverage on TAP-Vid-DAVIS. We therefore developed multi-hypothesis adaptive surface chaining (\mhac{}), which retains several candidate matches and an explicit missing state. After selection on disjoint development data, \mhac{} improved Average Jaccard over a position-only baseline by 0.1394 on a frozen ten-video holdout (95\% bootstrap interval $[0.1074,0.1689]$). Across 265 RoboTAP videos, the mean paired gain was 0.0610 ($[0.0510,0.0715]$), with 215 wins, one tie and 49 losses. These results support resource-bounded scene construction and an external tracking gain under the tested protocol. They do not establish metric scale, semantic identity or high-fidelity novel-view rendering.
\end{abstract}

\section{Introduction}

Video generators, monocular reconstruction models, and learned SLAM systems now provide strong but different views of a dynamic scene. Large video diffusion transformers synthesize plausible motion from text or an image \citep{wanteam2025wan}; feed-forward reconstructors infer time-varying geometry and trajectories from monocular RGB \citep{luo20264rc}; and learned dense SLAM estimates camera motion, maps, loop closures, and recovery state \citep{murai2025mast3rslam}. Combining these capabilities could provide a common route from an instruction, a reference image, or an observed video to a dynamic three-dimensional scene. In practice, the combination is fragile. Each component uses its own coordinate frame, scale, preprocessing, confidence convention, and memory profile, while generated frames must not be confused with physical observations.

The difficulty lies in coordinating outputs whose assumptions differ. A local reconstruction may have an arbitrary scale; a SLAM update may change the reference frame; a rejected match may leave a trajectory without support. Generated images also require different provenance from recorded observations. We therefore ask whether a common scene representation can preserve these distinctions, propagate coordinate changes consistently and remain within a single-GPU budget.

\begin{figure}[t]
    \centering
    \includegraphics[width=\textwidth]{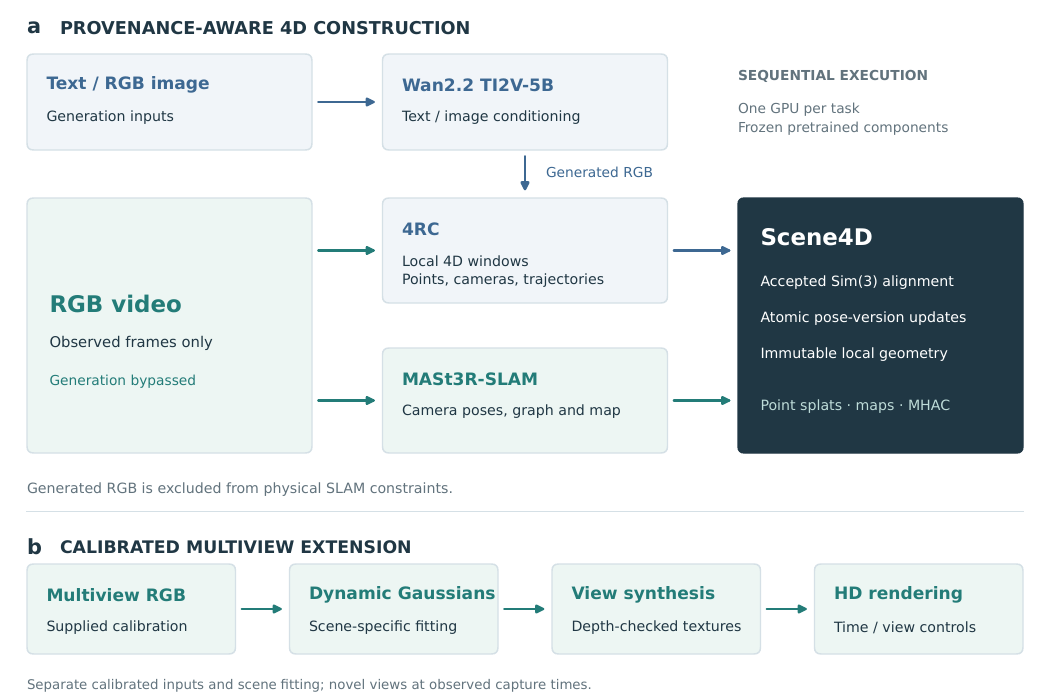}
    \caption{\textbf{System architecture and the separate calibrated rendering extension.} (a) Text and image inputs pass through video generation, while observed RGB enters reconstruction and SLAM directly. Frozen components execute sequentially on one GPU. Immutable local 4D windows are aligned through accepted Sim(3) transforms, and pose-version updates keep cameras, points and trajectories consistent. Generated RGB never constrains the physical SLAM map. (b) The high-resolution extension instead requires synchronized multiview RGB and supplied calibration. It fits each capture independently and renders observed times from new camera views; it is separate from the monocular evaluation.}
    \label{fig:overview}
\end{figure}

We study this problem under a strict operational contract: one process-intensive component at a time, one 40-GB-class GPU per task, RGB-only inference, short offline scenes, and relative rather than metric scale. Figure~\ref{fig:overview} summarizes the resulting system. Text and image inputs are first converted to RGB video; observed monocular videos bypass generation and additionally enter RGB-only SLAM. A windowed 4D reconstructor produces dense local point trajectories. Robust Sim(3) alignment maps accepted windows into a versioned scene frame, while rejected windows remain explicitly local. Motion evidence, equal-budget trajectory retention, and cross-window surface association operate on the exported representation instead of modifying pretrained networks.

Cross-window association exposes a second question: should temporal consistency be enforced by increasingly strict matches? Our predeclared TAP-Vid-DAVIS study gives a negative answer for the initial method. Aggregating geometry and colour over three overlap frames did not improve Average Jaccard (AJ) over a simpler one-frame rule and reduced the number of valid chains. This failure motivates \mhac{} (multi-hypothesis adaptive surface chaining), which preserves a bounded candidate lattice and trades geometric precision against coverage before query-specific decoding. We select one configuration on disjoint development data, pass a frozen ten-video holdout gate, and evaluate it once on all 265 RoboTAP videos under a hash-bound protocol.

The system provides:
\begin{enumerate}
    \item an executable three-input system that preserves provenance while producing explicit time-varying point geometry, camera trajectories, and RGB-only SLAM artifacts under a measured single-GPU contract;
    \item a versioned coordination layer in which local 4D arrays remain immutable and map corrections atomically update cameras, points, trajectories, and coordinate metadata;
    \item an evidence-based study of motion ranking, support-aware multi-frame surface correspondence, and fixed-budget trajectory retention on frozen real-video scenes; and
    \item a multi-hypothesis association procedure motivated by a failed strict-chain experiment and evaluated on a disjoint holdout and 265 external videos without post-test reselection.
\end{enumerate}

The evaluated profile uses point splats at relative scale and processes short scenes offline. Generated frames never enter the physical SLAM map. A separate calibrated multiview extension is shown in Section~\ref{sec:hd-extension}; it changes the input requirements and is not part of the monocular evaluation.

\section{Related work}

\paragraph{Video generation and 4D reconstruction.}
Diffusion and flow-based video models have scaled generation through latent representations and transformer backbones \citep{ho2022video,rombach2022ldm,peebles2023dit,lipman2023flow,wanteam2025wan}. Dynamic scene reconstruction instead seeks geometry and motion that remain coherent across time. Neural fields and Gaussian representations support dynamic novel-view synthesis \citep{wu2024fourDGS,yang2024deformablegs,lin2024gaussianflow}, while MoSca and Shape of Motion use motion scaffolds or low-dimensional motion bases to consolidate monocular cues \citep{lei2025mosca,wang2025shapeofmotion}. 4RC provides feed-forward dense 4D reconstruction and conditional temporal querying \citep{luo20264rc}. \method{} uses these advances as frozen components; its contribution lies in resource-bounded coordination, failure-preserving evaluation, and surface association rather than a new generative or reconstruction backbone.

\paragraph{Learned geometry and SLAM in dynamic scenes.}
Classical and learned SLAM systems estimate camera trajectories and persistent maps, with DROID-SLAM demonstrating recurrent dense bundle adjustment and ORB-SLAM3 providing a mature feature-based baseline \citep{teed2021droidslam,campos2021orbslam3}. DUSt3R and MASt3R recast matching and reconstruction in learned pointmap space \citep{wang2024dust3r,leroy2024mast3r}; MASt3R-SLAM builds tracking, local fusion, graph construction, loop closure, and global optimization on these priors \citep{murai2025mast3rslam}. Dynamic-scene systems remove uncertain regions or jointly represent static and moving content \citep{zheng2025wildgsslam,fan2026de4dslam}. We retain the full SLAM state machine for observed RGB but keep dynamic reconstruction outputs separate from map constraints. This avoids treating a generated or moving surface as evidence for the physical camera trajectory.

\paragraph{Long-range point and surface association.}
Tracking-any-point benchmarks measure both localization and occlusion across long videos \citep{doersch2022tapvid}. TAPIR, PIPs, CoTracker, and BootsTAP improve temporal refinement, joint tracking, or training data \citep{doersch2023tapir,harley2022pips,karaev2024cotracker,doersch2024bootstap}; RoboTAP demonstrates the relevance of persistent point tracks to robot imitation \citep{vecerik2024robotap}. Our cross-window problem differs from training a 2D tracker: each node is a dense 3D trajectory inside one reconstruction window, and edges must reconcile separate local coordinate systems. We use TAP-Vid metrics to quantify the resulting projected chains while keeping the graph query-independent until decoding.

\paragraph{Resource-bounded composition.}
Model compression and efficient adaptation reduce training or deployment cost, but peak memory in a composed system is also controlled by scheduling, activation lifetime, preprocessing, and persistent state. \method{} treats the resource envelope as an experimental variable. It records task-process-tree memory separately from unrelated device use, launches large components sequentially, validates every output before adoption, and reports the exact resolution, sequence length, and sampling interval to which each memory statement applies.

\section{Method}

\subsection{Problem formulation and scope}

The input is one of a text prompt $x^{\mathrm{text}}$, an RGB reference image $x^{\mathrm{img}}$, or a monocular RGB sequence $X=\{I_t\}_{t=0}^{T-1}$. The system returns a \scene{} artifact
\begin{equation}
\mathcal{S}=\big\{\mathcal{W}_k,\,\mathbf{A}_k,\,v_k,\,\mathcal{C},\,\mathcal{M}\big\}_{k=1}^{K},
\end{equation}
where $\mathcal{W}_k$ is an immutable local 4D window, $\mathbf{A}_k\in\mathrm{Sim}(3)$ maps that window to its declared scene frame, $v_k$ is a pose version, $\mathcal{C}$ contains per-frame cameras, and $\mathcal{M}$ is optional SLAM state for observed video. Text and image inputs first produce a generated RGB sequence $\hat X=G(x)$; they do not produce $\mathcal{M}$. All scales are relative.

The operational contract requires a single physical GPU per task and a sampled task-process-tree peak below $40{,}000{,}000{,}000$ bytes. Large neural components run in separate subprocesses and never remain resident concurrently. This contract is a measured property of a fixed profile, not a proof for arbitrary lengths or resolutions.

\subsection{Heterogeneous input routing}

For text and image modes, a frozen 5B Wan2.2 TI2V model produces 17 RGB frames at $832\times480$ using 12 sampling steps \citep{wanteam2025wan}. The text encoder executes on CPU and model weights are offloaded between stages. Image mode records the reference image hash and uses it only as the generator condition. Video mode decodes at most 60 frames and records source timestamps when available. Every frame manifest declares whether pixels are observed or generated. Only observed video enters SLAM.

The pipeline writes a durable stage record before starting each subprocess. A stage is accepted only when it exits successfully, resource logs are present, array shapes and finite values pass validation, and output hashes can be reloaded. Existing output directories are never overwritten. This design keeps implementation failures, preflight rejections, and completed artifacts distinguishable.

\subsection{Windowed explicit 4D representation}

The frozen 4RC backbone processes windows of $L=16$ frames with an overlap of four \citep{luo20264rc}. For valid time indices $t$ and dense source pixels $i$, a local window stores points $\mathbf{p}_{k,t,i}\in\mathbb{R}^3$, trajectory positions $\boldsymbol{\tau}_{k,t,i}\in\mathbb{R}^3$, camera poses $\mathbf{T}^{(k)}_{wc,t}$, intrinsics $\mathbf{K}^{(k)}_t$, colours, visibility states, and raw confidence scores. The identifier
\begin{equation}
q_{k,i}=(k\ll32)+i
\end{equation}
is stable inside a window but does not imply semantic identity across windows. Padded inputs needed by the backbone remain in the raw record; only valid frames enter the canonical artifact.

Local arrays are immutable after export. A scene-space point is materialized as
\begin{equation}
\mathbf{p}^{w}_{k,t,i}=s_k\mathbf{R}_k\mathbf{p}_{k,t,i}+\mathbf{t}_k,
\label{eq:sim3}
\end{equation}
where $\mathbf{A}_k=(s_k,\mathbf{R}_k,\mathbf{t}_k)$. Camera rotations receive $\mathbf{R}_k$ while camera centres receive the full similarity transform. This distinction prevents the scale component from corrupting orthonormal rotations.

\subsection{SLAM anchoring and atomic map updates}

Observed RGB runs a pinned MASt3R-SLAM instance before 4D reconstruction \citep{murai2025mast3rslam}. The SLAM branch exports timestamped camera poses, keyframes, pointmaps, graph edges, loop/relocalization events, and restartable backend state. Retrieval proposals are not reported as loop closures unless the backend accepts their constraints.

Each 4D window is aligned either to accepted SLAM keyframe geometry or to an already anchored overlapping window. Candidate static correspondences are split deterministically: 80\% fit a robust Sim(3), and 20\% measure held-out reprojection. Alignment is rejected if geometry is degenerate, consensus falls below 35\%, spatial coverage falls below 25\%, fewer than 20 held-out pairs have positive depth, or held-out median reprojection exceeds 3 pixels. Rejection preserves the window in a named local frame rather than inserting an identity transform and pretending global alignment.

When the SLAM graph changes, common keyframes induce a new window transform $\mathbf{A}'_k$. The update is committed only if every affected window passes centre-residual and rotation checks. Cameras, points, trajectories, and metadata then advance from $v_k$ to $v_k+1$ together. The source arrays and their hashes remain unchanged, so every global state can be audited against the same local reconstruction.

\subsection{Geometry--motion evidence and fixed-budget retention}

For trajectory $i$, we compute a world-coordinate residual score $r_i^{3d}$ after compensating for camera and accepted window transforms, and an image-motion residual $r_i^{2d}$ after a global image transform. Because their scales are incomparable and confidences are uncalibrated, we fuse ranks rather than multiply scores:
\begin{equation}
d_i=\tfrac{1}{2}\operatorname{rank}(r_i^{3d})+\tfrac{1}{2}\operatorname{rank}(r_i^{2d}).
\label{eq:rankfusion}
\end{equation}
The result is a category-agnostic motion diagnostic, not a semantic segmentation probability.

To retain at most $K$ trajectories, the budget sampler assigns half of the quota to spatial coverage and the remainder to high speed, turning, and visibility-transition events in proportions 0.20, 0.15, and 0.15. Every comparison uses identical $K$ and deterministic seeds. Sampling occurs after reconstruction, so it reduces stored and transmitted trajectory state but does not reduce the backbone's already incurred memory peak.

\subsection{From strict chains to multi-hypothesis adaptive association}

For adjacent windows $a$ and $b$, the initial P4 rule selected a single mutual nearest neighbour using three shared frames. This reduced false matches locally but could terminate an otherwise useful chain. \mhac{} instead constructs a query-independent directed lattice. For candidate edge $e=(i,j)$, the feature vector is
\begin{equation}
\mathbf{f}_e=\big[\bar r_e,\,r_e^{\max},\,\sigma(r_e),\,u_e,\,c_e,\,\rho_e^{a\rightarrow b},\,\rho_e^{b\rightarrow a},\,m_e\big],
\end{equation}
containing mean, maximum, and dispersion of reprojection error over four overlap frames; visible support fraction; normalized colour distance; forward and reverse neighbour rank; and the nearest-to-second-nearest margin. A fixed weighted score
\begin{equation}
C_e=\mathbf{w}^{\top}(\mathbf{f}_e\oslash\mathbf{s})
\label{eq:mhac-cost}
\end{equation}
ranks at most three outgoing candidates. The selected C02 configuration uses the eight frozen weights $(0.30,0.18,0.08,-0.12,0.16,0.06,0.06,-0.04)$, ordinary limits of 2-pixel mean reprojection, 4-pixel maximum reprojection, and 0.15 normalized colour distance, and no rescue edge.

The lattice is sealed before point queries or ground truth are loaded. Each benchmark query then performs bounded dynamic programming with a missing state. Query identities never interact, preventing one labelled track from informing another. Candidate selection used 30 calibration and 10 selection-validation videos from TAP-Vid RGB-Stacking; ten distinct development-holdout videos and all 265 RoboTAP videos were reserved under the protocol described in Section~\ref{sec:experimental-design}.

\section{Experimental design}
\label{sec:experimental-design}

\subsection{Frozen components and execution profile}

The main neural components contain 13,594,090,959 instantiated parameters: 4,999,787,712 in the video diffusion transformer, 5,680,910,336 in the text encoder, 704,688,668 in the VAE, 1,520,065,387 in 4RC, and 688,638,856 in MASt3R. The coordination layer adds no neural parameters and performs no end-to-end training. All weights, source revisions, and file hashes are pinned. The low-memory profile uses 17 generated frames or at most 60 observed frames, 16-frame reconstruction windows, four-frame overlap, and point-splat visualization.

GPU use is sampled every 50 ms. We report both the maximum memory attributable to the task process tree and total device use, and we flag any foreign compute process. This interval can miss shorter transients; all resource statements therefore retain the sampling qualifier. Runs launch only when the allowed device is otherwise idle.

\subsection{Datasets and split discipline}

The frozen real-video suite contains 16 short scenes drawn from TUM RGB-D, Bonn dynamic RGB-D, and DyCheck \citep{sturm2012tum,gao2022dycheck}. Ground-truth depth, pose, or motion references are isolated from inference and used only by the evaluator. Each scene contributes one independent unit; frames, pixels, tracks, and overlap folds are nested observations.

Long-range projected tracking uses TAP-Vid metrics at $256\times256$ \citep{doersch2022tapvid}. P4 used all 30 TAP-Vid-DAVIS videos once under a protocol frozen before payload access. P5 uses the 50 RGB-Stacking videos in a fixed 30/10/10 calibration, selection-validation, and development-holdout partition drawn before data access. TAP-Vid-DAVIS is prohibited from further tuning. The final external set is all 265 RoboTAP videos \citep{vecerik2024robotap}. Its payload was opened only after the selected method, code, analysis, holdout decision, and protocol hashes were frozen; every video remained in the denominator.

\subsection{Metrics and statistical units}

SLAM is summarized by absolute trajectory error (ATE), one-step translation RPE, and one-step rotation RPE when timestamp association meets the frozen threshold. Motion ranking uses ROC AUC and balanced accuracy against a geometric depth-change reference, not semantic labels. Surface association reports held-out reprojection error on correspondences excluded from fitting. Budget sampling reports event recall and spatial coverage at a common $K=8{,}192$.

Point tracking reports AJ, average position accuracy across 1, 2, 4, 8, and 16 pixels, and occlusion accuracy. Macro averages weight videos equally. Confidence intervals use 10,000 scene- or video-level percentile bootstrap resamples. Because the D25 analysis plan was written after its frozen test outputs were generated, those intervals are descriptive and no $p$-values are reported. P4 and P5 use their preregistered comparison directions and seeds.

\begin{table}[t]
\centering
\caption{\textbf{Evidence status in the completed frozen study.} ``Development'' indicates data used for design or deterministic selection and cannot support an external-generalization claim.}
\label{tab:evidence-status}
\small
\resizebox{\textwidth}{!}{%
\begin{tabular}{p{3.1cm}p{2.1cm}p{2.4cm}p{2.5cm}p{4.0cm}}
\toprule
Evidence block & Independent units & Status & Resource peak & Claim boundary \\
\midrule
Three-input acceptance & 9 runs + 3 fresh final runs & Complete & 17.14 GB & Executability of fixed short-scene profile; generation quality separate \\
Frozen real-video suite & 16 scenes & Complete & 16.27 GB & Point 4D reconstruction, RGB-only SLAM runtime, scene-level diagnostics \\
P4 TAP-Vid-DAVIS & 30 videos & Complete; primary hypothesis failed & 26.54 GB & External falsification of strict three-frame chain \\
P5 RGB-Stacking calibration & 30 videos & Complete; development & 16.27 GB & Candidate design only \\
P5 selection validation & 10 videos & Complete; development & 16.27 GB & Deterministic candidate selection only \\
P5 development holdout & 10 videos & Complete; gate passed & 16.27 GB & One-shot authorization of final external test \\
P5 RoboTAP & 265 videos & Complete; external gate passed & 16.27 GB & Bounded external robot-video tracking gain vs. B0 \\
\bottomrule
\end{tabular}
}
\end{table}

\clearpage
\section{Results}

\subsection{Calibrated multiview captures support high-resolution dynamic views}
\label{sec:hd-extension}

\begin{figure}[H]
\centering
\includegraphics[width=.93\textwidth]{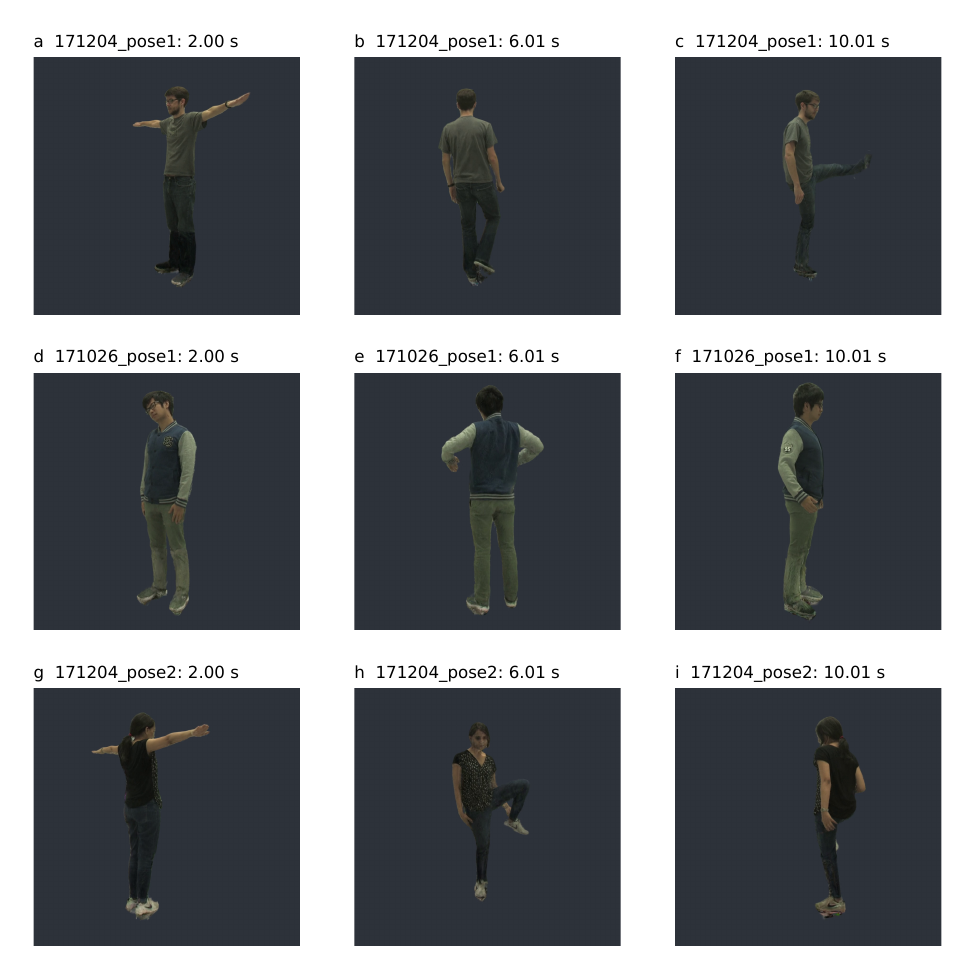}
\caption{\textbf{Calibrated multiview reconstruction, separate from the monocular experiments.} Rows show three separately fitted captures. Columns are frames 60, 180 and 300 of each 360-frame orbit video, at 2.002, 6.006 and 10.010 s; time and camera azimuth advance together. Panels use the same central crop of the native $1920\times1080$ frames, without additional colour adjustment or retouching. Training-view silhouette support reduces some detached splats, but shoe deformation and hand softness remain. Source recordings and calibration: CMU Panoptic Studio.}
\label{fig:hd-stills}
\end{figure}

The separate extension uses synchronized Panoptic Studio RGB recordings and supplied camera calibration \citep{joo2016panoptic}. It fits a dynamic Gaussian representation independently to each capture, then combines Gaussian appearance with geometry-checked textures from the fitting views. It does not use the monocular SLAM or 4RC pipeline. Figure~\ref{fig:hd-stills} shows uniformly spaced frames from three completed orbit videos.

The captures contain 360 observed frames each, spanning 12.012 s at $30000/1001$ fps. Their fitting/validation camera counts are 27/3, 24/3 and 26/3, respectively. The depicted times and subjects participated in scene fitting; the virtual camera positions are newly rendered views. Validation views were inspected during development. Thus, the images are neither unseen-scene inference nor an independent rendering benchmark. Residual footwear deformation and hand softness remain visible; the demonstration does not establish that perceptual quality is fully resolved.

\subsection{All three inputs completed within the measured single-GPU envelope}

The final acceptance run executed text, image, and observed-video inputs sequentially from fresh run directories. Text and image each generated 17 frames and produced two \scene{} submaps; the 60-frame observed video produced five submaps, a non-empty SLAM map, seven keyframes, and eight accepted graph edges. All exported scenes reloaded with array and hash validation. The maximum sampled task-process-tree footprint was 17,135,828,992 bytes (17.14 GB in decimal units), below both the 36-GB operating target and 40-GB hard contract. Across the earlier nine-run memory acceptance, every mode completed three times without generation or scene caches.

This result establishes execution of the fixed profile rather than quality at arbitrary duration. The generated examples use 17 frames and 12 steps; a separate 121-frame action-quality gate did not pass. The video branch uses at most 60 frames and relative scale. No experiment ran two large components concurrently.

Table~\ref{tab:three-inputs} reports the measured execution profile for each route.
% Generated from reports/phase3/d30_final_acceptance.json; do not edit numbers.
\begin{table}[H]
\centering
\caption{\textbf{Measured execution of the three input routes.} One fresh final
acceptance run per route, without generation or scene caches. Time is end-to-end
wall time in seconds; GPU memory is the sampled task-process-tree peak in decimal
GB. Pass denotes execution and exported-scene validation, not perceptual quality.}
\label{tab:three-inputs}
\small
\setlength{\tabcolsep}{6pt}
\begin{tabular}{lrrrrc}
\toprule
Input & Frames & Submaps & Time (s) & GPU (GB) & Execution \\
\midrule
Text & 17 & 2 & 262.32 & 17.12 & Pass \\
Image & 17 & 2 & 259.59 & 17.14 & Pass \\
Observed RGB & 60 & 5 & 155.03 & 16.27 & Pass \\
\bottomrule
\end{tabular}
\end{table}

\subsection{Frozen real videos expose both capability and failure}

\begin{figure}[t]
    \centering
    \includegraphics[width=\textwidth]{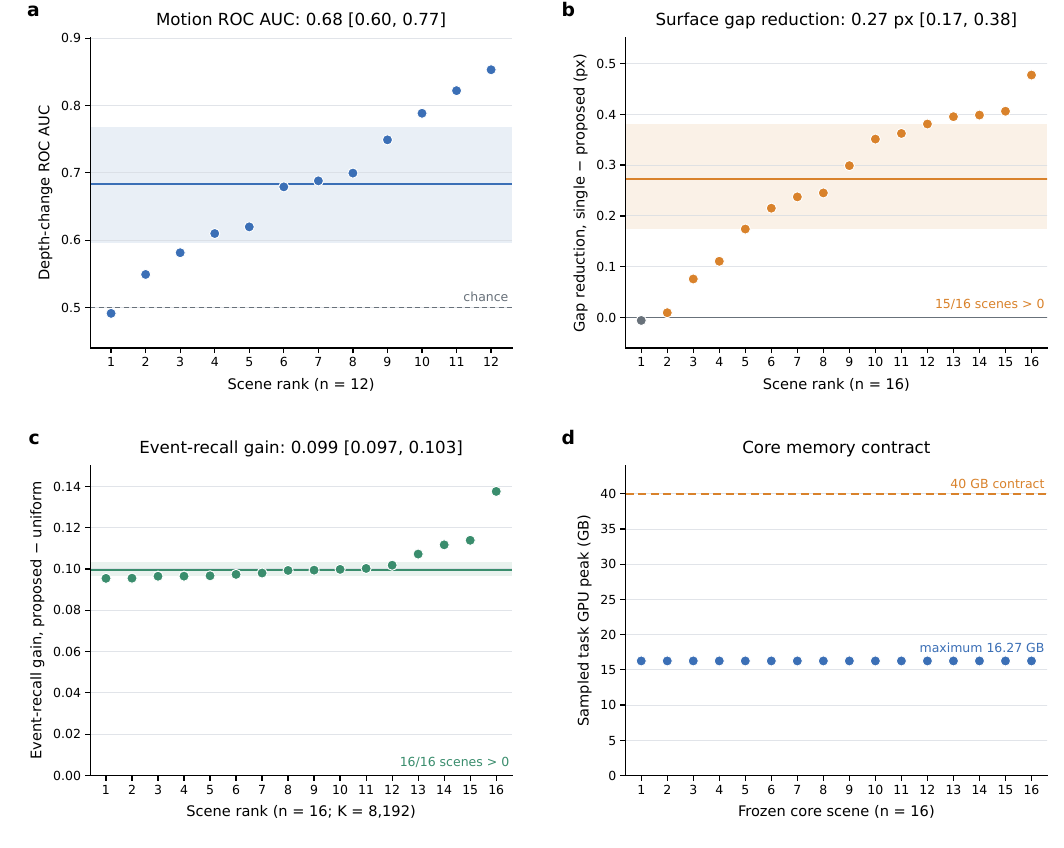}
    \caption{\textbf{Frozen real-video evaluation establishes a bounded resource and coordination result.} (a) Motion-ranking ROC AUC across 12 reference-equipped scenes; the dashed line is chance. (b) Paired reduction in held-out overlapping-surface error for the three-frame geometry--colour rule relative to a one-frame rule. (c) Paired event-recall gain for motion-diversity sampling relative to uniform sampling at $K=8{,}192$. (d) Sampled task-process-tree GPU peak for the 16 4D reconstructions relative to the 40-GB contract. Points are independent scenes; coloured horizontal marks and bands show scene medians and descriptive 10,000-resample percentile intervals where applicable.}
    \label{fig:heldout}
\end{figure}

All 16 frozen real-video scenes produced structurally valid 4RC outputs (Figure~\ref{fig:heldout}). Fifteen produced non-empty SLAM maps. The remaining scene, \texttt{dycheck\_wheel}, exited without a usable map and was retained as a failure. Of 12 scenes preregistered for pose evaluation, 11 met timestamp-association requirements. Their median ATE RMSE was 0.0170 m with a descriptive scene-bootstrap 95\% interval of $[0.0066,0.0428]$ m; median translation RPE was 0.0109 m $[0.0024,0.0211]$, and median mean rotation RPE was $0.718^\circ$ $[0.292^\circ,1.074^\circ]$. The nominal metric units follow the evaluation references and do not convert the complete output to a guaranteed metric-scale reconstruction.

The maximum task-process-tree GPU footprint across these 16 reconstructions was 16,265,510,912 bytes. The single SLAM failure and one unscorable pose case prevent a claim of universal robustness. They also demonstrate why success-rate accounting must remain separate from numeric metrics computed only on valid trajectories.

\subsection{Coordination modules improve two narrow diagnostics}

The equal-rank motion fusion achieved a median ROC AUC of 0.684 $[0.596,0.769]$ across 12 scenes and median balanced accuracy of 0.593 $[0.559,0.696]$ at the frozen 0.65 threshold. One scene had AUC 0.491, an explicit counterexample. Since the frozen test retained only the fused score, this result establishes absolute discrimination against the geometric reference but not superiority over either component score on test data.

Three-frame geometry--colour correspondence reduced scene-level median held-out reprojection error by 0.272 pixels $[0.174,0.381]$, or 27.1\% $[19.9\%,30.4\%]$, relative to the one-frame rule. Fifteen of 16 scenes improved. This endpoint is restricted to jointly visible surfaces in adjacent overlapping windows; it does not test object identity across long occlusions.

At the same $K=8{,}192$, motion-diversity sampling increased event recall by a median 0.0993 $[0.0967,0.1035]$ over uniform sampling and improved all 16 scenes. Median spatial-coverage ratio was 1.215 $[1.197,1.259]$. Because this sampler runs after reconstruction, the result supports storage-efficient trajectory retention but not lower neural-inference memory.

\subsection{External tracking falsifies the strict-chain hypothesis}

\begin{figure}[t]
    \centering
    \includegraphics[width=\textwidth]{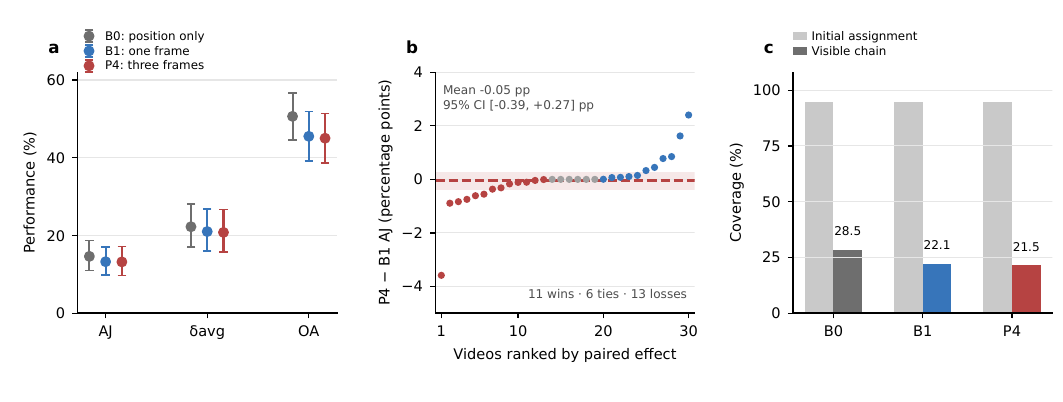}
    \caption{\textbf{The preregistered P4 external experiment rejects stricter matching as a sufficient route to long-range tracking.} (a) First-query AJ across all 30 TAP-Vid-DAVIS videos. (b) Paired video-level differences for the prespecified comparisons. (c) Query assignment and visible-chain coverage, showing that equal initial assignment does not prevent later chain loss. Intervals resample videos 10,000 times.}
    \label{fig:p4}
\end{figure}

P4 completed all 30 TAP-Vid-DAVIS videos and 1,999 frames without excluding low-performing cases (Figure~\ref{fig:p4}). In the preregistered first-query mode, the strict three-frame rule obtained AJ 13.21\%, compared with 13.26\% for the appearance-aware one-frame rule and 14.63\% for the position-only baseline. The paired P4--B1 difference was $-0.050$ percentage points with a video-bootstrap 95\% interval of $[-0.391,+0.275]$ percentage points. The P4--B0 difference was $-1.423$ points $[-2.280,-0.626]$.

The query assignment rate was common across variants, but visible-chain coverage fell from 28.53\% for B0 to 22.11\% for B1 and 21.47\% for P4. This diagnostic is consistent with premature pruning: the strict rule can improve local overlap precision yet lose complete projected tracks. It does not by itself prove that coverage is the only causal mechanism. The maximum task-process-tree GPU footprint was 26,543,652,864 bytes, and two contamination-triggered attempts were preserved separately before clean reruns.

\subsection{Frozen holdout and RoboTAP tests validate adaptive association}

\begin{figure}[t]
    \centering
    \includegraphics[width=\textwidth]{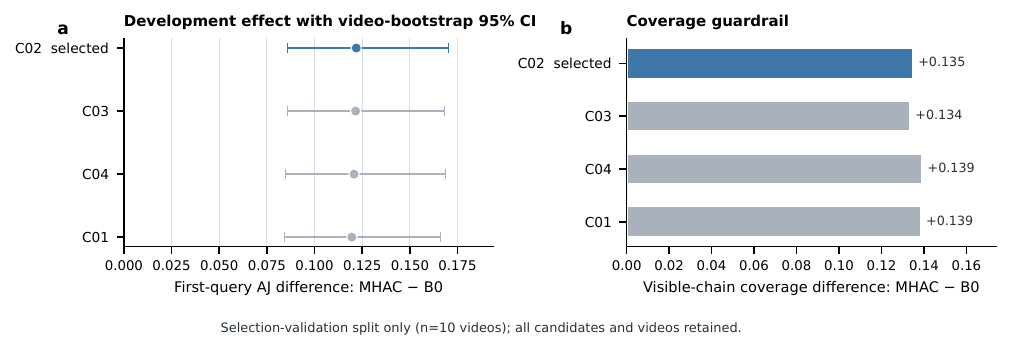}
    \caption{\textbf{P5 candidate selection on ten development videos.} (a) Paired first-query AJ improvement over the position-only B0 baseline; error bars are 10,000-resample video-bootstrap 95\% intervals. (b) visible-chain coverage improvement over B0. C02 was selected by the frozen lexicographic rule. These are development-selection results, not holdout or external-test evidence.}
    \label{fig:p5dev}
\end{figure}

The P4 result fixed the design objective for P5: retain a small set of plausible surface edges instead of tightening a single binary rule. Four candidates advanced from the 30-video calibration split and were evaluated without label access during reconstruction or graph construction on ten selection-validation videos. All four passed the prespecified coverage, occlusion, and candidate-count guardrails (Figure~\ref{fig:p5dev}).

The deterministic rule selected C02. Its first-query AJ was 0.4993, with a paired improvement over B0 of 0.1219 and a 95\% video-bootstrap interval of $[0.0856,0.1701]$; all ten videos improved. Visible-chain coverage was 0.6233, 0.1349 above B0, and occlusion accuracy was 0.7710. C02 and C03 differed by only 0.00035 AJ, so the evidence supports reproducible selection rather than statistical superiority of C02 over every alternative.

\begin{figure}[t]
    \centering
    \includegraphics[width=\textwidth]{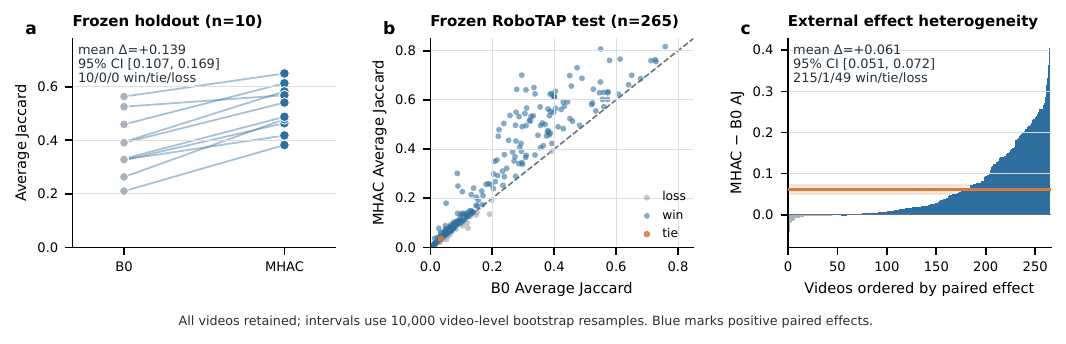}
    \caption{\textbf{Frozen confirmatory tests support a bounded external gain for \mhac{}.} (a) Every development-holdout video improves over B0. (b) Paired RoboTAP AJ for all 265 videos; the dashed identity line separates wins and losses. (c) Sorted paired effects expose both the positive mean and the retained negative cases; the orange line and band show the mean and its 10,000-resample video-bootstrap 95\% interval. No video was removed after evaluation.}
    \label{fig:p5confirm}
\end{figure}

The selection result was not used as confirmatory evidence. On the frozen ten-video development holdout, C02 improved AJ over B0 by 0.1394 with a video-bootstrap 95\% interval of $[0.1074,0.1689]$ and improved all ten videos (Figure~\ref{fig:p5confirm}). Macro visible-chain coverage increased by 0.2020. The mean-effect, win-count, coverage, and resource checks all passed, authorizing the hash-bound external protocol without candidate reselection.

All 265 RoboTAP videos then completed and remained in the aggregate. Mean first-query AJ increased from 0.1994 for B0 to 0.2604 for \mhac{}. The paired improvement was 0.0610 ($95\%$ interval $[0.0510,0.0715]$), with 215 wins, one tie, and 49 losses. Macro visible-chain coverage increased by 0.0685, and the maximum sampled task-process-tree footprint was 16,265,510,912 bytes (15.15 GiB). All four preregistered external checks passed. The 49 losses show material heterogeneity, so this result supports external robot-video point-tracking generalization against B0 under this protocol rather than universal superiority.

\subsection{The release is reproducible within its frozen profile}

The source release includes pinned third-party revisions, weight hashes, environment locks, a lightweight wheel, three command-line entry points, reference artifacts, and integrity manifests. A clean launcher installation, artifact reload, and all three entry modes passed; the historical final acceptance recorded 113 passing tests. Subsequent packaging checks recorded 119 passing tests without rerunning the full GPU pipeline. Full CUDA dependency installation from an empty machine image remains outside the completed evidence because the release uses prebuilt pinned environments. Licences remain those of the underlying components, including non-commercial restrictions where applicable.

\section{Discussion}

The results distinguish local correspondence accuracy from complete-track coverage. Three-frame matching lowered overlap reprojection error in the frozen scene suite, yet performed worse than the position-only baseline on TAP-Vid-DAVIS. The endpoints measure different properties: overlap error conditions on supported matches, whereas AJ also penalizes missing tracks. Retaining several candidates is useful when a locally plausible match should not determine the entire chain.

The RoboTAP gain is consistent with this interpretation. Visible-chain coverage increased by 6.85 percentage points alongside the AJ improvement. However, \mhac{} changes candidate retention, missing-state decisions and occlusion behaviour together, so the experiment does not isolate coverage as the sole cause. The 49 videos favouring B0 also show that the selected rule is not uniformly better. A future mechanism study would need fresh evaluation data and separate interventions on these decisions.

The systems result concerns coordinate consistency and resource use. Immutable local arrays preserve the reconstruction evidence, while versioned transforms keep map corrections consistent across cameras and trajectories. Sequential execution permits the frozen components to share one GPU. This arrangement is useful for offline inspection and reuse, but it does not remove uncertainty inherited from the components.

The evaluated system has several limits. Text and image runs establish execution for 17 frames; the separate 121-frame action-quality gate failed. Point splats do not establish high-fidelity novel views, and projected tracking does not establish semantic identity after occlusion. The external comparison is restricted to B0 on the frozen robot-video protocol, without a comprehensive comparison to specialized trackers. Resource measurements are sampled every 50 ms and may miss shorter peaks. The calibrated Gaussian extension provides a separate view-synthesis demonstration from dense multiview inputs, with unresolved footwear artefacts; it does not close these monocular evidence gaps.

\section{Conclusion}

\method{} produces versioned dynamic point scenes from three input routes and adds RGB-only SLAM for observed video. The tested short-scene profile ran within the measured single-GPU budget. The association experiments show why local geometric precision alone is insufficient: strict matching lost long-range coverage, while a bounded candidate lattice improved mean AJ over B0 on the frozen RoboTAP evaluation. The retained failures define the present scope of the result and the cases that require further study.

\section*{AI use statement}

AI tools were used solely to review experimental code and polish the language of the manuscript. The research ideas, experimental design and initial manuscript draft were developed independently by the authors. The authors take full responsibility for the accuracy and integrity of the final work.

\bibliography{references}

@inproceedings{rombach2022ldm,
  title={High-Resolution Image Synthesis with Latent Diffusion Models},
  author={Rombach, Robin and Blattmann, Andreas and Lorenz, Dominik and Esser, Patrick and Ommer, Bjorn},
  booktitle={Proceedings of the IEEE/CVF Conference on Computer Vision and Pattern Recognition},
  pages={10684--10695},
  year={2022},
  url={https://openaccess.thecvf.com/content/CVPR2022/html/Rombach_High-Resolution_Image_Synthesis_With_Latent_Diffusion_Models_CVPR_2022_paper.html}
}

@inproceedings{ho2022video,
  title={Video Diffusion Models},
  author={Ho, Jonathan and Salimans, Tim and Gritsenko, Alexey and Chan, William and Norouzi, Mohammad and Fleet, David J.},
  booktitle={Advances in Neural Information Processing Systems},
  volume={35},
  year={2022},
  url={https://proceedings.neurips.cc/paper/2022/hash/39235c56aef13fb05a6adc95eb9d8d66-Abstract-Conference.html}
}

@inproceedings{peebles2023dit,
  title={Scalable Diffusion Models with Transformers},
  author={Peebles, William and Xie, Saining},
  booktitle={Proceedings of the IEEE/CVF International Conference on Computer Vision},
  pages={4195--4205},
  year={2023},
  url={https://openaccess.thecvf.com/content/ICCV2023/html/Peebles_Scalable_Diffusion_Models_with_Transformers_ICCV_2023_paper.html}
}

@article{wanteam2025wan,
  title={Wan: Open and Advanced Large-Scale Video Generative Models},
  author={{WanTeam} and Wang, Ang and Ai, Baole and Wen, Bin and Mao, Chaojie and Xie, Chen-Wei and Chen, Di and Yu, Feiwu and Zhao, Haiming and others},
  journal={arXiv preprint arXiv:2503.20314},
  year={2025},
  url={https://arxiv.org/abs/2503.20314}
}

@inproceedings{lipman2023flow,
  title={Flow Matching for Generative Modeling},
  author={Lipman, Yaron and Chen, Ricky T. Q. and Ben-Hamu, Heli and Nickel, Maximilian and Le, Matt},
  booktitle={International Conference on Learning Representations},
  year={2023},
  url={https://openreview.net/forum?id=PqvMRDCJT9t}
}

@inproceedings{doersch2022tapvid,
  title={TAP-Vid: A Benchmark for Tracking Any Point in a Video},
  author={Doersch, Carl and Gupta, Ankush and Markeeva, Larisa and Recasens, Adria and Smaira, Lucas and Aytar, Yusuf and Carreira, Joao and Zisserman, Andrew and Yang, Yi},
  booktitle={Advances in Neural Information Processing Systems},
  volume={35},
  year={2022},
  url={https://proceedings.neurips.cc/paper_files/paper/2022/hash/58168e8a92994655d6da3939e7cc0918-Abstract.html}
}

@inproceedings{doersch2023tapir,
  title={TAPIR: Tracking Any Point with Per-Frame Initialization and Temporal Refinement},
  author={Doersch, Carl and Yang, Yi and Vecerik, Mel and Gokay, Dilara and Gupta, Ankush and Aytar, Yusuf and Carreira, Joao and Zisserman, Andrew},
  booktitle={Proceedings of the IEEE/CVF International Conference on Computer Vision},
  pages={10061--10072},
  year={2023},
  url={https://openaccess.thecvf.com/content/ICCV2023/html/Doersch_TAPIR_Tracking_Any_Point_with_Per-Frame_Initialization_and_Temporal_Refinement_ICCV_2023_paper.html}
}

@inproceedings{doersch2024bootstap,
  title={BootsTAP: Bootstrapped Training for Tracking-Any-Point},
  author={Doersch, Carl and Luc, Pauline and Yang, Yi and Gokay, Dilara and Koppula, Skanda and Gupta, Ankush and Heyward, Joseph and Rocco, Ignacio and Goroshin, Ross and Carreira, Joao and Zisserman, Andrew},
  booktitle={Proceedings of the Asian Conference on Computer Vision},
  pages={3257--3274},
  year={2024},
  url={https://openaccess.thecvf.com/content/ACCV2024/html/Doersch_BootsTAP_Bootstrapped_Training_for_Tracking-Any-Point_ACCV_2024_paper.html}
}

@inproceedings{harley2022pips,
  title={Particle Video Revisited: Tracking Through Occlusions Using Point Trajectories},
  author={Harley, Adam W. and Fang, Zhaoyuan and Fragkiadaki, Katerina},
  booktitle={European Conference on Computer Vision},
  pages={59--75},
  year={2022},
  publisher={Springer},
  url={https://arxiv.org/abs/2204.04153}
}

@inproceedings{karaev2024cotracker,
  title={CoTracker: It is Better to Track Together},
  author={Karaev, Nikita and Rocco, Ignacio and Graham, Benjamin and Neverova, Natalia and Vedaldi, Andrea and Rupprecht, Christian},
  booktitle={European Conference on Computer Vision},
  year={2024},
  url={https://arxiv.org/abs/2307.07635}
}

@inproceedings{luo20264rc,
  title={4RC: 4D Reconstruction via Conditional Querying Anytime and Anywhere},
  author={Luo, Yihang and Zhou, Shangchen and Lan, Yushi and Pan, Xingang and Loy, Chen Change},
  booktitle={Proceedings of the 43rd International Conference on Machine Learning},
  year={2026}
}

@inproceedings{murai2025mast3rslam,
  title={{MASt3R-SLAM}: Real-Time Dense {SLAM} with {3D} Reconstruction Priors},
  author={Murai, Riku and Dexheimer, Eric and Davison, Andrew J.},
  booktitle={Proceedings of the IEEE/CVF Conference on Computer Vision and Pattern Recognition},
  year={2025}
}

@inproceedings{wang2024dust3r,
  title={{DUSt3R}: Geometric 3D Vision Made Easy},
  author={Wang, Shuzhe and Leroy, Vincent and Cabon, Yohann and Chidlovskii, Boris and Revaud, Jerome},
  booktitle={Proceedings of the IEEE/CVF Conference on Computer Vision and Pattern Recognition},
  year={2024}
}

@article{leroy2024mast3r,
  title={Grounding Image Matching in 3D with {MASt3R}},
  author={Leroy, Vincent and Cabon, Yohann and Revaud, Jerome},
  journal={arXiv preprint arXiv:2406.09756},
  year={2024}
}

@inproceedings{teed2021droidslam,
  title={{DROID-SLAM}: Deep Visual {SLAM} for Monocular, Stereo, and {RGB-D} Cameras},
  author={Teed, Zachary and Deng, Jia},
  booktitle={Advances in Neural Information Processing Systems},
  volume={34},
  year={2021}
}

@article{campos2021orbslam3,
  title={{ORB-SLAM3}: An Accurate Open-Source Library for Visual, Visual--Inertial, and Multimap {SLAM}},
  author={Campos, Carlos and Elvira, Richard and Rodriguez, Juan J. Gomez and Montiel, Jose M. M. and Tardos, Juan D.},
  journal={IEEE Transactions on Robotics},
  volume={37},
  number={6},
  pages={1874--1890},
  year={2021}
}

@inproceedings{wu2024fourDGS,
  title={4D Gaussian Splatting for Real-Time Dynamic Scene Rendering},
  author={Wu, Guanjun and Yi, Taoran and Fang, Jiemin and Xie, Lingxi and Zhang, Xiaopeng and Wei, Wei and Liu, Wenyu and Tian, Qi and Wang, Xinggang},
  booktitle={Proceedings of the IEEE/CVF Conference on Computer Vision and Pattern Recognition},
  pages={20310--20320},
  year={2024}
}

@inproceedings{yang2024deformablegs,
  title={Deformable 3D Gaussians for High-Fidelity Monocular Dynamic Scene Reconstruction},
  author={Yang, Ziyi and Gao, Xinyu and Zhou, Wen and Jiao, Shaohui and Zhang, Yuqing and Jin, Xiaogang},
  booktitle={Proceedings of the IEEE/CVF Conference on Computer Vision and Pattern Recognition},
  pages={20331--20341},
  year={2024}
}

@inproceedings{lin2024gaussianflow,
  title={Gaussian-Flow: 4D Reconstruction with Dynamic 3D Gaussian Particle},
  author={Lin, Youtian and Dai, Zuozhuo and Zhu, Siyu and Yao, Yao},
  booktitle={Proceedings of the IEEE/CVF Conference on Computer Vision and Pattern Recognition},
  pages={21136--21145},
  year={2024}
}

@inproceedings{lei2025mosca,
  title={{MoSca}: Dynamic Gaussian Fusion from Casual Videos via 4D Motion Scaffolds},
  author={Lei, Jiahui and Weng, Yijia and Harley, Adam W. and Guibas, Leonidas and Daniilidis, Kostas},
  booktitle={Proceedings of the IEEE/CVF Conference on Computer Vision and Pattern Recognition},
  pages={6165--6177},
  year={2025}
}

@inproceedings{wang2025shapeofmotion,
  title={Shape of Motion: 4D Reconstruction from a Single Video},
  author={Wang, Qianqian and Ye, Vickie and Gao, Hang and Zeng, Weijia and Austin, Jake and Li, Zhengqi and Kanazawa, Angjoo},
  booktitle={Proceedings of the IEEE/CVF International Conference on Computer Vision},
  pages={9660--9672},
  year={2025}
}

@inproceedings{zheng2025wildgsslam,
  title={{WildGS-SLAM}: Monocular Gaussian Splatting {SLAM} in Dynamic Environments},
  author={Zheng, Jianhao and Zhu, Zihan and Bieri, Valentin and Pollefeys, Marc and Peng, Songyou and Armeni, Iro},
  booktitle={Proceedings of the IEEE/CVF Conference on Computer Vision and Pattern Recognition},
  pages={11461--11471},
  year={2025}
}

@inproceedings{fan2026de4dslam,
  title={{De4D-SLAM}: Gradient-Isolated Static-Dynamic Decoupling for Monocular {SLAM} in Dynamic Environments},
  author={Fan, Zhicheng and Wu, Zitong and Fan, Zhaoxing and Zhang, Xiao and Hou, Biao and Ren, Bo},
  booktitle={Proceedings of the 43rd International Conference on Machine Learning},
  series={Proceedings of Machine Learning Research},
  volume={306},
  pages={28883--28903},
  year={2026}
}

@inproceedings{vecerik2024robotap,
  title={{RoboTAP}: Tracking Arbitrary Points for Few-Shot Visual Imitation},
  author={Vecerik, Mel and Doersch, Carl and Yang, Yi and Davchev, Todor and Aytar, Yusuf and Zhou, Guangyao and Hadsell, Raia and Agapito, Lourdes and Scholz, Jon},
  booktitle={IEEE International Conference on Robotics and Automation},
  year={2024}
}

@inproceedings{sturm2012tum,
  title={A Benchmark for the Evaluation of {RGB-D} {SLAM} Systems},
  author={Sturm, J{ü}rgen and Engelhard, Nikolas and Endres, Felix and Burgard, Wolfram and Cremers, Daniel},
  booktitle={IEEE/RSJ International Conference on Intelligent Robots and Systems},
  pages={573--580},
  year={2012}
}

@inproceedings{gao2022dycheck,
  title={Monocular Dynamic View Synthesis: A Reality Check},
  author={Gao, Hang and Li, Ruilong and Tulsiani, Shubham and Russell, Bryan and Kanazawa, Angjoo},
  booktitle={Advances in Neural Information Processing Systems},
  volume={35},
  year={2022}
}

@article{joo2016panoptic,
  title={Panoptic Studio: A Massively Multiview System for Social Interaction Capture},
  author={Joo, Hanbyul and Simon, Tomas and Li, Xulong and Liu, Hao and Tan, Lei and Gui, Lin and Banerjee, Sean and Godisart, Timothy and Nabbe, Bart and Matthews, Iain and Kanade, Takeo and Nobuhara, Shohei and Sheikh, Yaser},
  journal={arXiv preprint arXiv:1612.03153},
  year={2016},
  url={https://arxiv.org/abs/1612.03153}
}
\bibliographystyle{iclr2027_conference}

\appendix

\section{Artifact and claim map}

Table~\ref{tab:artifact-map} maps each principal statement to its authoritative artifact. Development records are kept separate from frozen-test and external records.

\begin{table}[h]
\centering
\caption{Authoritative evidence artifacts for the reported results.}
\label{tab:artifact-map}
\small
\resizebox{\textwidth}{!}{%
\begin{tabular}{p{4.0cm}p{7.2cm}p{3.3cm}}
\toprule
Claim & Artifact & Evidence class \\
\midrule
Three-input final acceptance & \path{reports/phase3/d30_final_acceptance.json} & Final engineering acceptance \\
Frozen 16-scene metrics & \path{reports/phase3/d25_statistics.json} and case CSV & Frozen test; descriptive inference \\
P4 external negative result & \path{reports/phase4/p4_tapvid_aggregate.json} & Preregistered external test \\
P5 candidate selection & \path{reports/phase5/p5_mhac_selection_validation.json} & Development selection \\
P5 frozen holdout & \path{reports/phase5/p5_mhac_development_holdout.json} & One-shot development holdout \\
P5 RoboTAP external result & \path{reports/phase5/p5_robotap_external_aggregate.json} & Frozen external test \\
P5 completion audit & \path{reports/phase5/p5_final_completion_audit.json} & Protocol and execution audit \\
Model count and provenance & \path{models_manifest.json} & Hash-verified implementation record \\
\bottomrule
\end{tabular}
}
\end{table}

\section{Scene4D field semantics}

Each local submap stores timestamps, frame identifiers, $\mathbf{T}_{wc}$, intrinsics, image size, dense points, point confidence, RGB, dense trajectories, trajectory identifiers, colour, trajectory confidence, and ternary visibility. Confidence values are raw model scores and are not interpreted as calibrated probabilities. Visibility denotes projection/depth support, not ground-truth occlusion. A static-probability value of $-1$ denotes unknown. These conventions are validated on load and preserved in the scene manifest.

\section{Failure and missingness policy}

A task can end as completed, failed, blocked by a frozen prerequisite, or contaminated by an unrelated GPU process. Failures remain in task counts. A contaminated attempt may be rerun only under a rule frozen before its outcome is known, and both attempts remain recorded. Numeric summaries do not impute missing trajectories or empty maps. Predeclared denominator units are scenes or videos, never frames or pixels.

\section{Confirmatory evaluation audit}

The final P5 result was admitted only after the frozen completion audit verified the following conditions:
\begin{enumerate}
    \item the ten-video holdout passed its four-part gate before external payload access;
    \item the final protocol bound the selected method, holdout report, input inventory, analysis seed, and hashes;
    \item all 265 admitted RoboTAP units completed and remained in the aggregate;
    \item inference and labels were isolated, task processes used one GPU each, and the sampled peak remained below 40 GB; and
    \item the reported bootstrap interval reproduces under the frozen seed and no candidate was reselected after external evaluation.
\end{enumerate}
All 18 machine-readable completion checks passed. The audit records retries, raw-output pruning, proxy state, resource logs, and per-video report hashes; it does not convert the external result into evidence for claims outside the specified protocol.

\section{Exact working-profile configuration}

The current released profile uses one allowed physical GPU, no concurrent large-model process, $832\times480$ generation, 17 frames, 12 sampling steps, CPU text encoding, model offload, 60 maximum observed frames, 10 frames per second for un-timestamped inputs, 16-frame reconstruction windows, four-frame overlap, 512-pixel reconstruction size, and point-splat rendering. The hard task limit is $40{,}000{,}000{,}000$ bytes and the operating target is $36{,}000{,}000{,}000$ bytes.

\end{document}